\documentclass[10pt,twocolumn,letterpaper]{article}

\usepackage{cvpr}              
\usepackage{multirow}
\usepackage{booktabs}
\usepackage{xcolor}
\usepackage{makecell}
\usepackage{arydshln}
\definecolor{largegray}{gray}{0.40}
\newcommand{\largescale}[1]{\textcolor{largegray}{#1}}
\definecolor{cvprblue}{rgb}{0.21,0.49,0.74}
\usepackage[pagebackref,breaklinks,colorlinks,allcolors=cvprblue]{hyperref}

\title{CascadeJEPA: Enhancing 3D Geometric Representations via \\ Spatially Coupled Cascaded Prediction}

\author{
Kailai Zhou$^{1}$ \quad
Ziqi Chen$^{2}$ \quad
Xiaoning Xu$^{2}$ \quad
Yuefeng Wu$^{2}$ \quad
Zixuan Fu$^{1}$ \quad
Zijun Lin$^{1}$ \\ 
Zhihao Li$^{1}$ \quad
Youwei Pang$^{1}$ \quad
Xiaoqi Zhao$^{1}$ \quad
Xun Cao$^{2}$ \quad
Bihan Wen$^{1}$ \\[6pt] 
$^{1}$Nanyang Technological University \qquad
$^{2}$Nanjing University
}

\begin{document}
\maketitle
\begin{abstract}
Joint-Embedding Predictive Architectures (JEPAs) have emerged as a promising paradigm for predictive visual representation learning. However, our analysis of I-JEPA reveals a growing discrepancy between semantic and geometric learning: improved semantic recognition during pretraining is accompanied by degraded 3D geometric representations. We hypothesize that this partly stems from predicting highly abstract latent representations with incomplete spatial context, favoring semantic regularities over fine-grained geometric information. To address this issue, we introduce CascadeJEPA, a spatially coupled cascaded prediction framework that promotes richer patch-level interactions through full-grid context encoding and  cross-attention between prediction stages, while largely preserving the original I-JEPA architecture. Extensive experiments demonstrate substantial improvements over I-JEPA across multiple 3D geometric tasks while maintaining strong semantic recognition performance.  Furthermore, its advantages extend to future feature forecasting and world model planning, highlighting the potential of enhanced geometric representations to support predictive world modeling.

\end{abstract}    
\section{Introduction}
\label{sec:intro}
Understanding the visual world requires not only recognizing objects and  semantic relationships, but also perceiving their shapes, spatial arrangements, and 3D structure. These complementary capabilities are essential for visual systems to understand and interact with the physical world \cite{goodale1992separate}.

\begin{figure}[t]  
  \centering
    \includegraphics[width=1.0\linewidth]{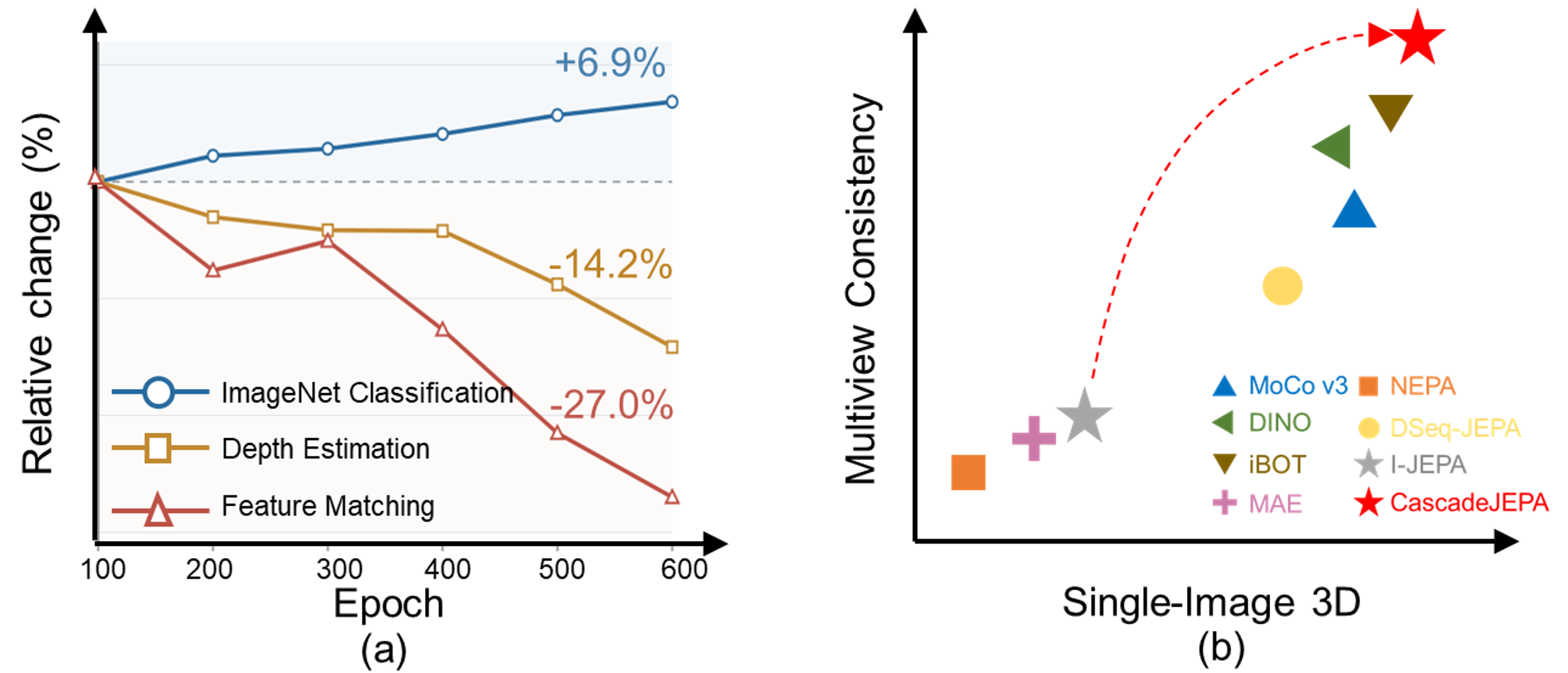} 
    \vspace{-1.5em}
    \caption{(a) Analysis of I-JEPA~\cite{assran2023ijepa}   across different pretraining stages reveals a discrepancy between semantic abstraction and 3D geometric preservation. (b)  CascadeJEPA achieves significant improvements over I-JEPA on fundamental 3D geometric tasks. }  
    \label{fig:intro}    
     \vspace{-1em}
\end{figure}

Joint-Embedding Predictive Architectures (JEPAs) offer a promising framework toward this goal by predicting target representations directly in latent space rather than reconstructing raw pixels~\cite{lecun2022path}. However, our evaluation shows that most existing JEPA-based models~\cite{assran2023ijepa,xu2025nepa,he2026dseqjepa,kuhn2026levjepa,bardes2024vjepa,assran2025vjepa2}  struggle on 3D geometric tasks. In particular, probing I-JEPA~\cite{assran2023ijepa} across pretraining stages shows that the steady improvement in ImageNet~\cite{russakovsky2015imagenet} classification accuracy does not translate into better geometric representations. As illustrated in Fig.~\ref{fig:intro}(a), continued pretraining overall degrades feature matching accuracy and depth estimation~\cite{jampani2023navi} performance, with deeper-layer features exhibiting particularly pronounced losses in spatial correspondence.

These observations reveal a growing discrepancy between semantic abstraction and geometric representation during I-JEPA pretraining. We hypothesize that this discrepancy partly arises from the conventional I-JEPA prediction paradigm, which relies on incomplete spatial context to predict highly abstract latent representations.  Specifically, conventional I-JEPA encodes only a small subset of visible patches, limiting spatial feature interactions to these patches while excluding masked positions from the contextual feature grid.
Meanwhile, its single-stage prediction directly targets highly abstract final-layer representations, potentially favoring predictable semantic regularities over fine-grained spatial variations. Together, these characteristics may encourage semantic abstraction without sufficiently preserving the local geometric distinctions. 



Meanwhile, as shown in Fig.~\ref{fig:intro}(b), methods emphasizing invariance-driven learning through cross-view representation alignment, such as DINO~\cite{caron2021emerging}, iBOT ~\cite{zhou2022ibot}, and MoCo v3~\cite{chen2021empirical}, generally exhibit stronger 3D geometric representations than prediction-driven methods~\cite{he2022masked,assran2023ijepa}. Notably, these methods align representations across augmented views while largely preserving their spatial organization, suggesting that spatially structured feature interactions may benefit geometric representation learning. This observation motivates us to reconsider how spatial relationships are modeled during latent prediction. Can we enhance JEPA's 3D geometric representations while preserving its semantic learning capabilities and architectural simplicity?

To this end,  we introduce CascadeJEPA, a simple extension of I-JEPA that enhances geometric representations through spatially coupled cascaded prediction. Rather than relying on a single prediction stage,  CascadeJEPA organizes latent prediction into two stages connected through cross-attention. The first stage retains the full spatial grid to learn spatially structured contextual representations, which subsequently condition fine-grained latent prediction in the second stage.  This unidirectional coupling leverages full-grid spatial context to promote richer patch-level interactions, facilitating the learning of geometrically informative representations.
Our main contributions are as follows:

\begin{itemize}
    \item We identify a growing discrepancy between semantic and geometric representation learning during I-JEPA pretraining, revealing that improved semantic recognition can be accompanied by degraded 3D geometric understanding.

    \item We propose CascadeJEPA, a simple spatially coupled cascaded prediction framework that promotes richer patch-level interactions to enhance geometric representation learning, while largely preserving I-JEPA architecture.

    \item With minimal architectural modifications, CascadeJEPA achieves substantial improvements across multiple 3D geometric tasks while maintaining competitive semantic performance. Its benefits extend to future feature forecasting and world model planning, highlighting the value of geometric  representations for predictive world models.
\end{itemize}



\section{Related Work}
\label{sec:relatedwork}
\subsection{Visual Representation Learning}

Learning generalizable visual representations is a fundamental objective of computer vision. With the emergence of Vision Transformers (ViTs)~\cite{dosovitskiy2020image}, early approaches~\cite{touvron2021training,touvron2022deit} primarily relied on supervised learning, whose dependence on large-scale annotated datasets limits scalability. Self-supervised learning (SSL) addresses this limitation by learning visual representations directly from unlabeled images. ViT-based SSL methods can be broadly characterized by two learning objectives: invariance-driven alignment and prediction-driven learning. Invariance-driven methods encourage consistent representations across different views of an image. MoCo v3~\cite{chen2021empirical} aligns augmented views through contrastive learning, while DINO~\cite{caron2021emerging} and iBOT~\cite{zhou2022ibot} employ teacher--student self-distillation to learn transferable visual features. These approaches typically rely on strong data augmentations and carefully designed training strategies. 

Prediction-driven methods learn representations by predicting missing visual information from observed context, often with simpler design and light augmentation. MAE~\cite{he2022masked} reconstructs masked pixels from visible patches, encouraging the encoder to capture contextual information. Joint-embedding predictive architectures~\cite{lecun2022path,assran2023ijepa} shift prediction from pixel space to representation space, learning to predict latent features of unobserved regions. By avoiding explicit pixel-level reconstruction, JEPAs provide a promising foundation for world models capable of understanding and predicting physical environments~\cite{lecun2022path}.

\subsection{Joint-Embedding Predictive Architectures}

I-JEPA~\cite{assran2023ijepa} introduces latent prediction for image self-supervised learning, where a context encoder processes visible image regions and a predictor estimates the representations of masked target regions generated by a momentum target encoder. Building upon this framework, subsequent studies explore different prediction strategies and learning objectives. DSeq-JEPA~\cite{he2026dseqjepa} replaces parallel region prediction with attention-guided sequential prediction, progressively predicting visual regions according to their discriminative importance. Relatedly, NEPA~\cite{xu2025nepa} adopts an autoregressive formulation that predicts subsequent patch embeddings from preceding ones using a single Transformer. LeJEPA~\cite{balestriero2025lejepa} introduces a theoretically grounded objective that regularizes representations toward an isotropic Gaussian distribution, enabling stable training without relying on heuristics such as stop-gradient or EMA teachers. 

Beyond static images, JEPA has also been extended to video representation learning to capture spatio-temporal dynamics.  V-JEPA~\cite{bardes2024vjepa} extends latent prediction to masked spatio-temporal regions in videos. V-JEPA~2~\cite{assran2025vjepa2} further scales video pretraining and introduces an action-conditioned predictor for physical prediction and robotic planning. More recently, V-JEPA~2.1~\cite{murlabadia2026vjepa21} improves spatially grounded representations through dense predictive supervision over both visible and masked tokens, together with self-supervision at multiple encoder layers.

Despite these advances, our analysis reveals that existing JEPA-based models generally exhibit weaker 3D geometric representations than invariance-driven methods. This observation motivates us to revisit conventional latent prediction and explore more effective strategies for preserving geometric information within the JEPA architecture.
\section{Method}
\label{sec:method}

\subsection{3D Geometric Representation Analysis}
\label{sec:geometry_analysis}

How geometric representations of I-JEPA~\cite{assran2023ijepa}  evolve during pretraining remains underexplored. We investigate this question through two fundamental 3D geometric tasks: feature matching for cross-view consistency and monocular depth estimation for single-image 3D structure.

\textbf{Semantic improvement accompanies geometric degradation.}
As shown in Fig.~\ref{fig:intro}(a), ImageNet classification accuracy steadily improves as I-JEPA pretraining progresses from 100 to 600 epochs, whereas performance on depth estimation and feature matching~\cite{jampani2023navi} degrades by 14.2\% and 27.0\%, respectively. This contrasting trend indicates that prolonged pretraining enhanced semantic discrimination is accompanied by a decline in geometric representation quality, highlighting a growing discrepancy between semantic and geometric learning.

\begin{figure}[t]  
  \centering
    \includegraphics[width=1.0\linewidth]{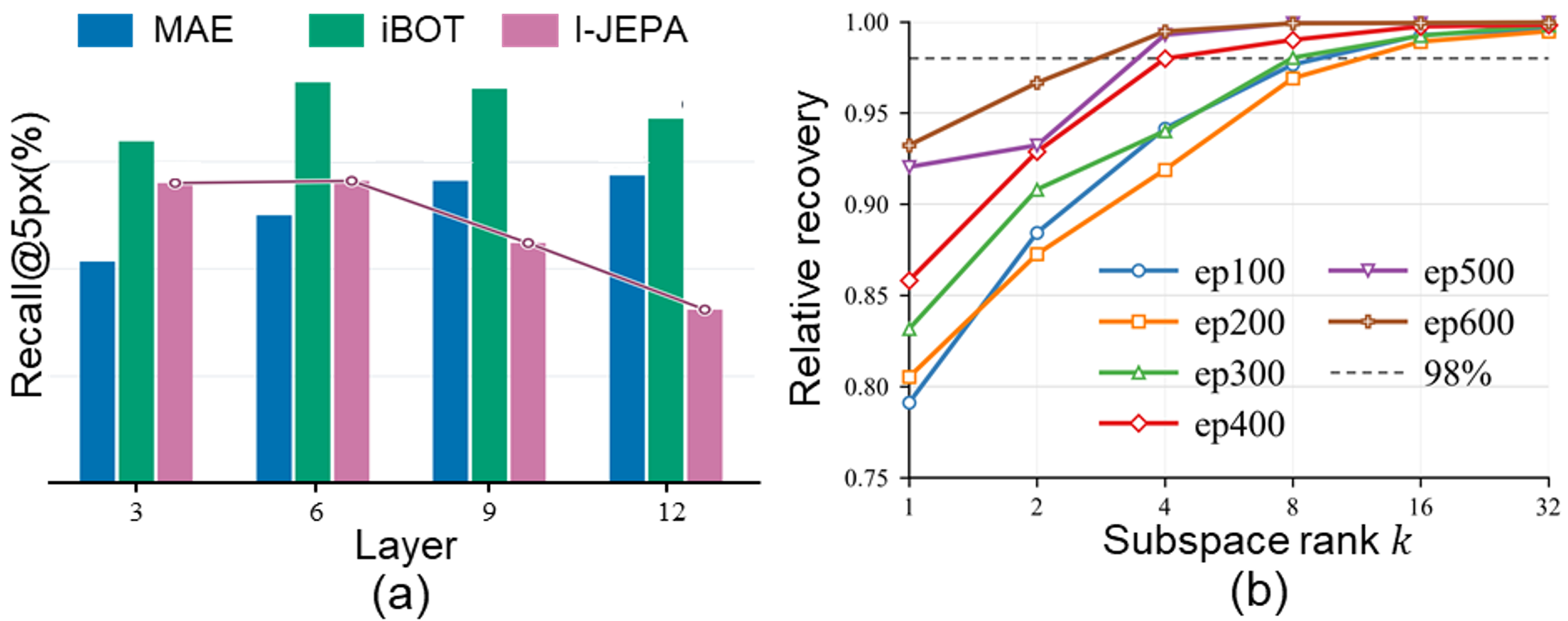} 
    \vspace{-1.5em}
    \caption{(a) Layer-wise feature matching on NAVI shows that geometric correspondence in I-JEPA peaks at intermediate layers and degrades substantially in deeper layers.
(b) Subspace intervention on depth estimation reveals increasing concentration of depth-relevant information as pretraining progresses.}  
    \label{fig:geo_analysis}    
    \vspace{-1.0em}
\end{figure}

\textbf{Diminishing geometric correspondence in deeper layers.}
To examine how correspondence changes with Transformer depth, we extract features from layers 3, 6, 9, and 12 of MAE, iBOT, and I-JEPA and evaluate them using Recall@5px on the NAVI~\cite{jampani2023navi} dataset. As shown in Fig.~\ref{fig:geo_analysis}(a), I-JEPA maintains comparable correspondence performance in early layers but exhibits a substantial decline in deeper layers. In contrast, MAE improves with depth, while iBOT remains relatively stable. These results suggest that I-JEPA preserves geometric correspondence in intermediate layers but progressively loses fine-grained spatial distinctions in deeper layers.


\textbf{Progressive concentration of geometric information.}
Based on the subspace-intervention analysis~\cite{zhou2026understanding},  we further investigate how linearly decodable depth information is distributed across feature directions. Let $s_e(k)$ denote the depth score obtained using a rank-$k$ intervention on the fused multi-layer representation at epoch $e$. We define the relative recovery with respect to the full representation as $r_e(k)=s_e(k)/s_e(\mathrm{full})$. Fig.~\ref{fig:geo_analysis}(b) shows that at epoch 100, rank $4$ recovers only $94.16\%$ of the full-representation score, while rank $16$ is the smallest rank that exceeds the $98\%$ threshold, reaching $99.27\%$. At epoch 600, rank $1$ already recovers $93.24\%$, and rank $4$ reaches $99.47\%$. Increasing the rank further from $4$ to $32$ provides only a marginal gain of $0.50$ percentage points, yielding $99.97\%$ recovery. However, this earlier saturation does not correspond to stronger depth estimation: the absolute full-representation score decreases by $14.2\%$ from epoch 100 to epoch 600, as shown in Fig.~\ref{fig:intro}(a). Meanwhile, from epoch 100 to 600, the minimum rank required for $98\%$ recovery decreases stepwise from $16$ to $4$. These results indicate that linearly decodable depth information becomes increasingly concentrated in a lower-dimensional subspace as pretraining progresses, potentially suggesting a tendency toward geometric over-compression.

\begin{figure*}[t]  
  \centering
    \includegraphics[width=0.95\linewidth]{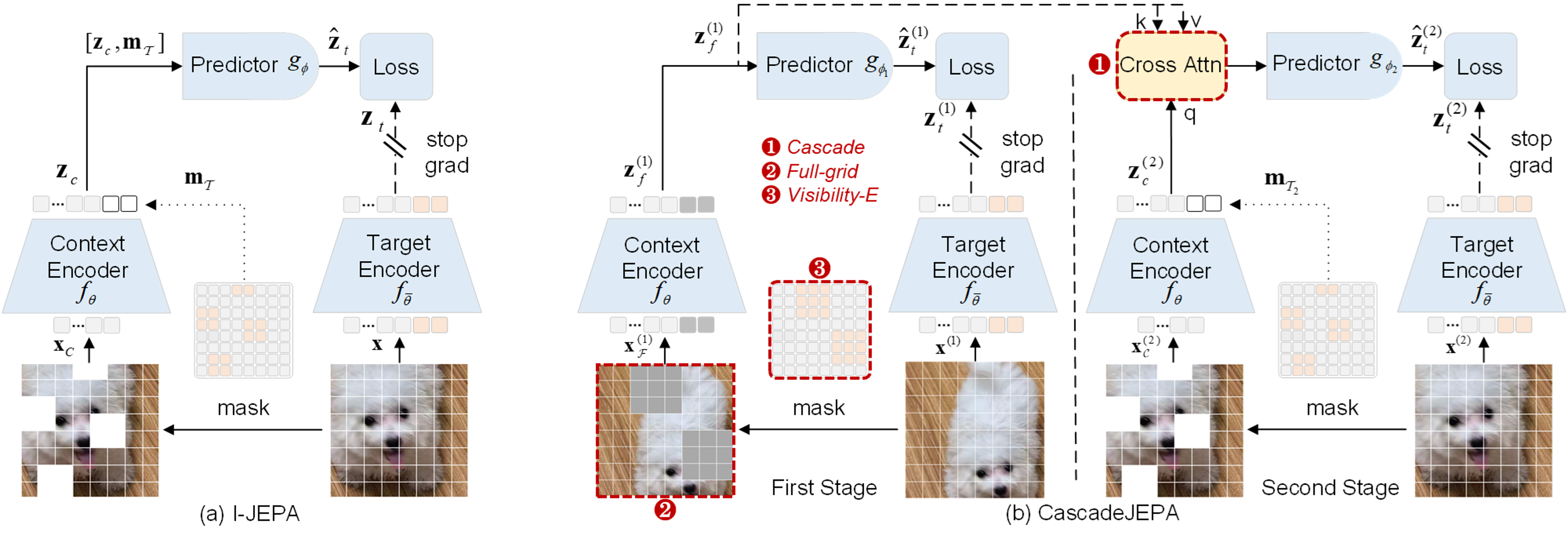} 
    \vspace{-1em}
    \caption{(a) I-JEPA performs single-stage latent prediction using contextual representations encoded from visible patches. (b) CascadeJEPA introduces spatially coupled cascaded prediction, where the first stage learns spatially structured contextual representations through full-grid encoding, and the second stage uses them to guide latent prediction via cross-attention. This unidirectional coupling promotes richer patch-level interactions while largely preserving the original I-JEPA architecture.
 }
    \label{fig:method}  
    \vspace{-1em}
\end{figure*}

Together, these findings suggest that I-JEPA learns some meaningful geometric representations, but may progressively compress them into coarse-grained structures as pretraining progresses and network depth increases.

\subsection{CascadeJEPA}
\subsubsection{Limitations of I-JEPA Latent Prediction}

The observed degradation in geometric representations motivates us to investigate two potential limitations of conventional I-JEPA latent prediction:

\textbf{1) Incomplete Spatial Context.} I-JEPA encodes only visible patches, leaving masked positions absent from the contextual feature grid. Consequently, spatial feature interactions are restricted to the visible subset, potentially limiting the modeling of fine-grained geometric relationships across the full spatial grid.

\textbf{2) Single-stage Prediction of Abstract Latent Features.} I-JEPA directly predicts final-layer target representations from contextual features extracted from incomplete visible patches. As representations become increasingly abstract in deeper layers, fine-grained geometric distinctions may become less pronounced. As shown in Fig.~\ref{fig:geo_analysis}(a), geometric correspondence degrades substantially in deeper layers. In particular, patches belonging to the same object may share similar semantic representations despite differences in depth and local structure. Consequently, single-stage latent prediction based on limited spatial context may favor semantic regularities without sufficiently capturing these geometric distinctions.

Inspired by human vision, where spatial and object information are closely integrated to support coherent scene perception~\cite{connor2017integration}, we seek to incorporate complementary spatial information into abstract latent prediction to better preserve fine-grained geometric details.  To this end, we extend single-stage latent prediction into a spatially coupled cascaded framework, where the first stage provides complementary spatial context to guide second-stage prediction.

\subsubsection{Preliminaries: I-JEPA}

As illustrated in Fig.~\ref{fig:method}(a), I-JEPA \cite{assran2023ijepa} learns visual representations by predicting the latent features of masked image regions from visible context. Given an input image $\mathbf{x}\in\mathbb{R}^{H\times W\times 3}$, the image is first divided into non-overlapping patches. A subset of patches is selected as the visible context $\mathcal{C}$, while the sampled target blocks to be predicted are denoted by $\mathcal{T}$. I-JEPA employs two encoders: a context encoder $f_\theta$ and a target encoder $f_{\bar{\theta}}$. The context encoder processes only the visible patches to extract contextual representations, while the target encoder processes the full image to generate target representations at the masked positions. This process can be formulated as:

\begin{equation}
\mathbf{z}_c = f_\theta(\mathbf{x}_{\mathcal C}), \qquad
\mathbf{z}_t = \operatorname{sg}\left([f_{\bar\theta}(\mathbf{x})]_{\mathcal T}\right),
\end{equation}

\noindent where $\operatorname{sg}(\cdot)$ denotes the stop-gradient operation. The target encoder shares the same architecture as the context encoder, but its parameters are updated through an exponential moving average (EMA) of the context encoder.

Given the contextual features $\mathbf{z}_c$, a lightweight predictor $g_\phi$ inserts learnable mask tokens at the masked spatial positions and predicts their corresponding latent representations:
$
\hat{\mathbf{z}}_t = g_\phi([\mathbf{z}_c,\mathbf{m}_{\mathcal T}]),
$
where $\mathbf{m}_{\mathcal T}$ denotes learnable mask tokens associated with the target positions, and $[\cdot,\cdot]$ denotes concatenation along the token dimension.
The model is trained by minimizing the squared L2 distance between the predicted representations and the corresponding target features:

\begin{equation}
\mathcal{L}_{\mathrm{I\text{-}JEPA}} =
\frac{1}{|\mathcal T|}
\sum_{i\in\mathcal T}
\left\|
\hat{\mathbf{z}}_{t,i}
-
\mathbf{z}_{t,i}
\right\|_2^2.
\end{equation}

\begin{table*}[t]
\centering
\caption{\textbf{Geometric and semantic correspondence.}
We report correspondence recall on NAVI
and ScanNet at different pixel thresholds,
and on SPair-71k across viewpoint differences $d$.
\textbf{Bold} denotes the best prediction-driven results pretrained on ImageNet-1K.}
\label{tab:correspondence}
\vspace{-2mm}
\footnotesize
\setlength{\tabcolsep}{5.5pt}
\renewcommand{\arraystretch}{1.12}

\begin{tabular*}{\textwidth}{
lccc
rrr
rrr
rrrr
}
\toprule
\multirow{2}{*}{Method} &
\multirow{2}{*}{Arch.} &
\multirow{2}{*}{\shortstack{Pre-train \\epochs}}  &
\multirow{2}{*}{\shortstack{Pre-train\\data}} &
\multicolumn{3}{c}{NAVI 2D Recall $\uparrow$} &
\multicolumn{3}{c}{ScanNet 2D Recall $\uparrow$} &
\multicolumn{4}{c}{SPair-71k Recall $\uparrow$} \\
\cmidrule(lr){5-7}
\cmidrule(lr){8-10}
\cmidrule(lr){11-14}
& & & &
5px & 25px & 50px &
5px & 10px & 20px &
$d=0$ & $d=1$ & $d=2$ & All \\
\midrule

\multicolumn{14}{l}{
\textit{Invariance-driven methods}
} \\
\addlinespace[2pt]

MoCo v3~\cite{chen2021empirical} &
ViT-B/16 & 300 & ImageNet-1K &
2.58 & 23.31 & 40.00 &
25.03 & 39.31 & 51.00 &
21.40 & 17.30 & 18.46 & 19.06 \\

DINO~\cite{caron2021emerging} &
ViT-B/16 & 400 & ImageNet-1K &
3.48 & 22.46 & 37.50 &
24.38 & 34.22 & 45.47 &
30.77 & 25.92 & 26.77 & 28.07 \\

iBOT~\cite{zhou2022ibot} &
ViT-B/16 & 400 & ImageNet-1K &
3.39 & 23.71 & 40.02 &
20.04 & 29.45 & 41.07 &
40.89 & 32.97 & 35.65 & 37.28 \\

\midrule
\multicolumn{14}{l}{
\textit{Prediction-driven methods}
} \\
\addlinespace[2pt]
\largescale{LeVJEPA~\cite{kuhn2026levjepa}} &
\largescale{ViT-L/16} & \largescale{--} &
\largescale{VideoMix} &
\largescale{1.97} & \largescale{12.33} &
\largescale{22.64} & \largescale{14.24} &
\largescale{21.53} & \largescale{32.17} &
\largescale{10.45} & \largescale{7.82} &
\largescale{7.59} & \largescale{9.28} \\
\largescale{V-JEPA 1~\cite{bardes2024vjepa}} &
\largescale{ViT-L/16} & \largescale{--} &
\largescale{VideoMix2M} &
\largescale{2.75} & \largescale{18.71} &
\largescale{32.53} & \largescale{6.74} &
\largescale{11.99} & \largescale{21.05} &
\largescale{32.93} & \largescale{18.63} &
\largescale{16.60} & \largescale{26.49} \\
\largescale{V-JEPA 2~\cite{assran2025vjepa2}} &
\largescale{ViT-L/16} & \largescale{--} &
\largescale{VideoMix22M} &
\largescale{2.36} & \largescale{15.91} &
\largescale{28.07} & \largescale{7.78} &
\largescale{13.21} & \largescale{22.46} &
\largescale{33.01} & \largescale{25.74} &
\largescale{27.20} & \largescale{29.46} \\
\largescale{V-JEPA 2.1~\cite{murlabadia2026vjepa21}} &
\largescale{ViT-G/16} & \largescale{--} &
\largescale{VisionMix163M} &
\largescale{5.11} & \largescale{31.37} &
\largescale{46.63} & \largescale{30.90} &
\largescale{41.00} & \largescale{51.11} &
\largescale{60.89} & \largescale{46.59} &
\largescale{46.48} & \largescale{54.23} \\
\addlinespace[2pt]
MAE~\cite{he2022masked} &
ViT-B/16 & 1600 & ImageNet-1K &
2.73 & 16.08 & 27.09 &
6.64 & 10.31 & 18.42 &
12.37 & 8.71 & 6.76 & 10.34 \\
NEPA~\cite{xu2025nepa} &
ViT-B/14 & 1600 & ImageNet-1K &
0.92 & 6.05 & 14.21 &
6.59 & 13.13 & 24.27 &
4.82 & 4.66 & 4.87 & 4.78 \\
DSeq-JEPA~\cite{he2026dseqjepa} &
ViT-B/16 & 600 & ImageNet-1K &
2.28 & 15.94 & 29.64 &
5.49 & 10.76 & 21.06 &
27.39 & 21.50 & 22.08 & 24.74 \\
I-JEPA~\cite{assran2023ijepa} &
ViT-B/16 & 600 & ImageNet-1K &
1.59 & 12.08 & 24.75 &
1.52 & 3.90 & 10.92 &
22.14 & 14.84 & 14.82 & 18.50 \\
\textbf{CascadeJEPA} &
ViT-B/16 & 300 & ImageNet-1K &
\textbf{4.26} &
\textbf{25.17} &
\textbf{39.42} &
\textbf{26.11} &
\textbf{34.05} &
\textbf{42.91} &
\textbf{45.67} &
\textbf{32.39} &
\textbf{33.59} &
\textbf{40.10} \\
\bottomrule
\end{tabular*}
\vspace{-3mm}
\end{table*}

\subsubsection{From I-JEPA to CascadeJEPA}

Building upon I-JEPA, we introduce CascadeJEPA, a spatially coupled cascaded prediction framework designed to enhance 3D geometric representation learning. As illustrated in Fig.~\ref{fig:method}(b), CascadeJEPA extends the original single-stage prediction into two cascaded stages with minimal architectural modifications. Our design consists of three key modifications: 1) cascaded architecture, 2)  full-grid context encoding, and 3) visibility-enhanced spatial coupling.

\textbf{1) Cascaded Architecture.} Unlike I-JEPA, which predicts masked latent features in a single stage, CascadeJEPA connects two I-JEPA stages through cross-attention. Specifically, the two stages receive different crops of the same input image, while sharing the weights of the context encoder and target encoder. The first stage extracts contextual representations, which are subsequently incorporated into the second-stage predictor through cross-attention. This allows the second stage to leverage additional spatial contextual information from the first stage, rather than relying solely on its own visible regions. Importantly, our cascaded design preserves the original I-JEPA architecture, introducing only a cross-attention module.

\textbf{2) Full-grid Context Encoding.}  As discussed in Sec.~\ref{sec:intro}, invariance-driven methods~\cite{chen2021empirical,caron2021emerging,zhou2022ibot} generally exhibit stronger geometric representations than prediction-driven methods~\cite{he2022masked,assran2023ijepa}. One relevant distinction lies in how spatial information is organized in the encoder. While methods such as MoCo v3 and DINO process complete patch grids of augmented image views, MAE and I-JEPA encode only visible patches, leaving masked positions absent from the contextual feature grid. This design limits explicit feature interactions across the full spatial grid. Motivated by this observation, we retain the complete spatial grid in the first-stage context encoder by replacing masked patch embeddings with zeros before adding positional embeddings. By retaining spatially indexed tokens at masked locations, the encoder facilitates contextual interactions between visible and masked positions, encouraging the learning of geometry-relevant spatial relationships.




\textbf{3) Visibility-Enhanced Spatial Coupling.} To facilitate the learning of fine-grained spatial relationships, we increase the proportion of visible patches in the first stage to provide richer spatial context. Specifically, we reduce the number of masked target blocks from 4 to 2, while slightly increasing the target-block scale to maintain a balance between spatial context preservation and semantic abstraction. The resulting full-grid contextual representations are then incorporated into the second-stage predictor through cross-attention, establishing unidirectional spatial coupling. 

Overall, CascadeJEPA retains the simplicity of the original I-JEPA  while enabling the second stage to leverage richer spatial context through cross-attention, thereby facilitating patch-level interactions and promoting the preservation of fine-grained geometric information.

\subsubsection{  CascadeJEPA Formulation}

Given an input image $\mathbf{x}$, CascadeJEPA samples two distinct crops $\mathbf{x}^{(1)}$ and $\mathbf{x}^{(2)}$ following the original I-JEPA cropping strategy for the two  stages, respectively. Positional embeddings are defined independently within the coordinate of each resized crop.
Both stages share the context encoder $f_\theta$ and
the EMA target encoder $f_{\bar\theta}$, while adopting
different masking configurations.
Specifically, the first stage retains the complete spatial
grid by replacing masked patch embeddings with zeros,
whereas the second stage follows I-JEPA and processes
only visible patches. The corresponding context and
target representations are obtained as:

\begin{equation}
\begin{aligned}
\mathbf{z}_f^{(1)} &= f_\theta(\mathbf{x}_{\mathcal F}^{(1)}),
\qquad
\mathbf{z}_t^{(1)} = \operatorname{sg}\left(
[f_{\bar\theta}(\mathbf{x}^{(1)})]_{\mathcal T_1}
\right),\\
\mathbf{z}_c^{(2)} &= f_\theta(\mathbf{x}_{\mathcal C}^{(2)}),
\qquad
\mathbf{z}_t^{(2)} = \operatorname{sg}\left(
[f_{\bar\theta}(\mathbf{x}^{(2)})]_{\mathcal T_2}
\right).
\end{aligned}
\end{equation}

\noindent where $\mathbf{x}_{\mathcal F}^{(1)}$ denotes the full-grid
input with zero-masked patch embeddings added with
positional embeddings, while $\mathbf{x}_{\mathcal C}^{(2)}$
represents the visible patch tokens in the second stage.

We introduce a cross-attention module to incorporate the first-stage spatial context into
the second-stage prediction. Specifically, the second-stage
contextual features together with the mask tokens serve as queries, while the first-stage
full-grid representations provide keys and values:

\begin{equation}
\tilde{\mathbf{z}}_{c}^{(2)} =[\mathbf{z}_c^{(2)},\mathbf{m}_{\mathcal T_2}] +
\operatorname{CrossAttn}
\left(
[\mathbf{z}_c^{(2)},\mathbf{m}_{\mathcal T_2}],
\mathbf{z}_f^{(1)},
\mathbf{z}_f^{(1)}
\right).
\end{equation}

The spatially conditioned features
$\tilde{\mathbf{z}}_{c}^{(2)}$ are fed into the
second-stage predictor to obtain
$\hat{\mathbf{z}}_t^{(2)} =
g_{\phi_2}\left(\tilde{\mathbf{z}}_{c}^{(2)}\right)$,
while the first-stage predictor independently produces
$\hat{\mathbf{z}}_t^{(1)} =
g_{\phi_1}\left(\mathbf{z}_f^{(1)}\right)$.
Finally, the two stages are jointly optimized by
minimizing their respective latent prediction losses:

\begin{equation}
\begin{aligned}
\mathcal{L}
=\ \frac{\lambda_1}{|\mathcal T_1|}
\sum_{i\in\mathcal T_1}
\left\|
\hat{\mathbf z}_{t,i}^{(1)}
-
\mathbf z_{t,i}^{(1)}
\right\|_2^2
+\frac{\lambda_2}{|\mathcal T_2|}
\sum_{i\in\mathcal T_2}
\left\|
\hat{\mathbf z}_{t,i}^{(2)}
-
\mathbf z_{t,i}^{(2)}
\right\|_2^2.
\end{aligned}
\end{equation}
where $\lambda_1$ and $\lambda_2$ denote the loss weights for the first and second stages, respectively. We assign equal weights to both stages, i.e., $\lambda_1=\lambda_2=0.5$. Further implementation details are provided in the supplementary material.

\begin{figure*}[t]  
  \centering
    \includegraphics[width=1.0\linewidth]{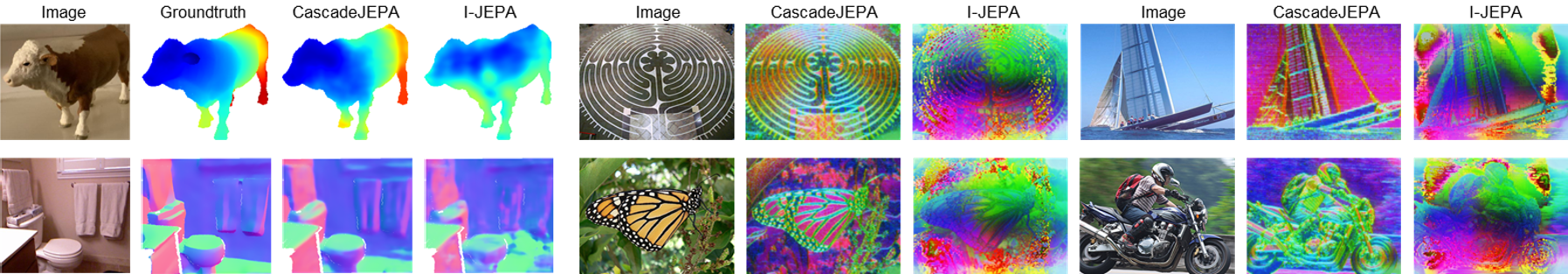} 
    \vspace{-2.0em}
        \caption{Comparison of I-JEPA and CascadeJEPA on depth estimation, surface normal estimation, and dense feature visualization.}  
    \label{fig:vis_task} 
    \vspace{-1.0em}
\end{figure*}
\section{Experiments}
\label{sec:experiments}

\subsection{Quantitative and Qualitative Evaluation}
\label{sec:quantitative}

\textbf{Feature Correspondence.}
We evaluate the spatial discriminability and consistency of learned representations through geometric and semantic correspondence tasks. For geometric correspondence, we consider object-centric matching on NAVI~\cite{jampani2023navi} and scene-level matching on ScanNet~\cite{dai2017scannet}, reporting 2D correspondence recall at pixel thresholds of 5/25/50 and 5/10/20, respectively. For semantic correspondence, we evaluate cross-instance keypoint matching on SPair-71k~\cite{min2019spair} across three viewpoint difficulty levels ($d=0,1,2$). Based on ProbeMid~\cite{chen2025probing}, we extract dense features from final ViT layers and establish correspondences without task-specific training.

 Tab.~\ref{tab:correspondence} shows invariance-driven methods generally outperform prediction-driven approaches in feature correspondence. Among the latter, MAE~\cite{he2022masked} achieves stronger geometric matching than I-JEPA~\cite{assran2023ijepa}, suggesting that pixel-level reconstruction may better preserve fine-grained spatial details. Moreover, V-JEPA~\cite{bardes2024vjepa} and V-JEPA 2~\cite{assran2025vjepa2} exhibit limited geometric correspondence despite larger backbones and extensive video pretraining, consistent with our observations in Sec.~\ref{sec:geometry_analysis}. V-JEPA 2.1~\cite{murlabadia2026vjepa21} substantially improves correspondence through dense predictive supervision and multi-level training, but employs a much larger backbone and VisionMix163M pretraining data.

\begin{table}[t]
\centering
\caption{Evaluation on surface normal and depth estimation.}
\label{tab:normal_depth}
\vspace{-2mm}
\footnotesize
\setlength{\tabcolsep}{3pt}
\renewcommand{\arraystretch}{1.12}
\begin{tabular*}{\columnwidth}{@{\extracolsep{\fill}}lrrrrrr@{}}
\toprule
\multirow{2}{*}{Method} &
\multicolumn{3}{c}{NYUv2 $\uparrow$} &
\multicolumn{3}{c}{NAVI $\uparrow$} \\
\cmidrule(lr){2-4}
\cmidrule(l){5-7}
& $11.25^\circ$ & $22.5^\circ$ & $30^\circ$
& $\delta_1$ & $\delta_2$ & $\delta_3$ \\
\midrule
\multicolumn{7}{l}{\textit{Invariance-driven methods}} \\
\addlinespace[2pt]
MoCo v3~\cite{chen2021empirical}
& 48.76 & 69.77 & 77.40 & 55.15 & 79.49 & 89.27 \\
DINO~\cite{caron2021emerging}
& 48.83 & 68.99 & 76.70 & 54.61 & 79.05 & 88.93 \\
iBOT~\cite{zhou2022ibot}
& 51.75 & 71.91 & 79.12 & 55.83 & 80.40 & 89.92 \\
\midrule
\multicolumn{7}{l}{\textit{Prediction-driven methods}} \\
\addlinespace[2pt]
\largescale{LeVJEPA (L)~\cite{kuhn2026levjepa}}
& \largescale{45.41}
& \largescale{66.90}
& \largescale{74.91}
& \largescale{46.76}
& \largescale{72.61}
& \largescale{84.88} \\
\largescale{V-JEPA 1 (L)~\cite{bardes2024vjepa}}
& \largescale{52.86}
& \largescale{70.25}
& \largescale{76.94}
& \largescale{51.51}
& \largescale{76.72}
& \largescale{87.79} \\
\largescale{V-JEPA 2 (L)~\cite{assran2025vjepa2}}
& \largescale{49.85}
& \largescale{68.02}
& \largescale{75.21}
& \largescale{50.66}
& \largescale{75.71}
& \largescale{86.82} \\
\largescale{V-JEPA 2.1 (G)~\cite{murlabadia2026vjepa21}}
& \largescale{64.17}
& \largescale{80.50}
& \largescale{86.11}
& \largescale{71.57}
& \largescale{89.18}
& \largescale{94.64} \\
\addlinespace[2pt]
MAE~\cite{he2022masked}
& 43.37 & 65.13 & 73.56 & 38.29 & 64.75 & 79.40 \\
NEPA~\cite{xu2025nepa}
& 32.33 & 56.00 & 66.63 & 33.85 & 59.49 & 75.73 \\
DSeq-JEPA~\cite{he2026dseqjepa}
& 49.49 & 69.50 & 76.83 & 51.83 & 76.88 & 87.65 \\
I-JEPA~\cite{assran2023ijepa}
& 44.24 & 66.35 & 74.52 & 43.50 & 69.84 & 83.22 \\
\textbf{CascadeJEPA} & \textbf{51.52} & \textbf{71.50} & \textbf{78.67} & \textbf{56.31} & \textbf{79.87} & \textbf{89.51} \\
\bottomrule
\end{tabular*}
\vspace{-3mm}
\end{table}

In contrast, CascadeJEPA achieves substantial improvements across three correspondence datasets while retaining the same backbone and pretraining dataset as I-JEPA, with only lightweight cross-stage attention introduced. Specifically, it improves NAVI Recall@25px by 13.09 points (12.08 to 25.17), ScanNet Recall@10px by 30.15 points (3.90 to 34.05), and overall SPair-71k recall by 21.60 points (18.50 to 40.10), surpassing all compared invariance-driven methods in overall SPair-71k recall. These consistent gains demonstrate the effectiveness of spatially coupled cascaded prediction in enhancing  discriminative representations for both geometric and semantic correspondence.

\begin{table}[t]
\centering\caption{Comparison of fine-tuning results on ImageNet-1K.}
\label{tab:semantic_comparison}
\vspace{-0.5em}
\footnotesize
\setlength{\tabcolsep}{3.8pt}
\renewcommand{\arraystretch}{1.12}
\begin{tabular}{llccc}
    \toprule
    Method & Arch.  & Pre-train epochs & Pre-train data  & Top-1 \\
    \midrule
    MAE~\cite{he2022masked}
    & ViT-B/16 & 300 & ImageNet-1K & 83.10 \\
    MAE~\cite{he2022masked}
    & ViT-B/16 & 1600  & ImageNet-1K & 83.48  \\
    I-JEPA~\cite{assran2023ijepa}
    & ViT-B/16 & 600 & ImageNet-1K  &  82.52\\
    \textbf{CascadeJEPA}
    & ViT-B/16 & 300  & ImageNet-1K & \textbf{83.50}  \\
    \bottomrule
\end{tabular}
\vspace{-1em}
\end{table}

\textbf{Surface Normal Estimation.}
To assess fine-grained geometric understanding, we evaluate surface normal estimation on NYUv2~\cite{silberman2012indoor}. Following Probe3D~\cite{el2024probing}, we adopt a DPT-style prediction head that aggregates multi-level features from frozen ViT backbones, training only the prediction head. For video-based models, we repeat the same image across temporal frames for consistent single-image evaluation. We report the percentage of pixels with angular errors below $11.25^\circ$, $22.5^\circ$, and $30^\circ$.


As shown in Tab.~\ref{tab:normal_depth}, prediction-driven methods generally exhibit weaker surface normal estimation. Notably, DSeq-JEPA~\cite{he2026dseqjepa} achieves competitive performance, suggesting that improved prediction strategies can benefit geometric representation learning. CascadeJEPA consistently outperforms I-JEPA across all angular thresholds, with gains of 7.28, 5.15, and 4.15 percentage points, respectively. The largest improvement at the strictest $11.25^\circ$ threshold highlights its advantage in capturing fine-grained surface geometry. Moreover, CascadeJEPA achieves the best performance among ImageNet-1K-pretrained prediction-driven methods, even surpassing several V-JEPA variants pretrained on substantially larger video datasets, demonstrating the effectiveness of spatially coupled cascaded prediction for local geometric representation learning.

\begin{table*}[t]
\centering
\caption{Ablation studies on the three key components of spatially coupled cascaded prediction in CascadeJEPA.}
\vspace{-1.0em}
\label{tab:component_ablation}
\footnotesize
\setlength{\tabcolsep}{5.8pt}
\renewcommand{\arraystretch}{1.12}

\begin{tabular}{lccccccccccccc}
    \toprule
    \multirow{2}{*}{Method}
    & \multicolumn{3}{c}{ Components}
    & \multicolumn{3}{c}{Depth Estimation}
    & \multicolumn{3}{c}{Geometric Matching}
    & \multicolumn{3}{c}{Semantic Matching} \\
    \cmidrule(lr){2-4}
    \cmidrule(lr){5-7}
    \cmidrule(lr){8-10}
    \cmidrule(l){11-13}
    & Cross-Attn. & Full-grid & Visibility-E
    & $\delta_1$ & $\delta_2$ & $\delta_3$
    & 5px & 25px & 50px
    & $d=0$ & $d=1$ & $d=2$ \\
    \midrule

    I-JEPA$_{epoch600}$
    & & &  &
        43.50 & 69.84 & 83.22 & 1.59
    & 12.08 & 24.75 & 22.14
    & 14.84 & 14.82  \\

    I-JEPA$_{epoch300}$
    & & &  &
    48.57 & 74.56 & 86.38 & 2.25
    & 15.32 & 28.52 & 25.65
    & 20.85 & 21.81\\
    
    Cascade Baseline
    & $\checkmark$ & &  &
    51.10 & 76.01 & 87.03 & 2.29
    & 16.06 & 29.63 & 29.67
    & 23.13 & 23.40  \\

    + Full-grid
    & $\checkmark$ & $\checkmark$ & &
    52.74 & 77.70 & 88.22 & 2.55
    & 17.41 & 31.37 & 34.40
    & 27.18 & 28.11  \\

    + Visibility-E
    & $\checkmark$ &   & 
    $\checkmark$  &55.12 &79.28 &88.96 &
    2.59& 18.09& 32.60& 38.21 & 28.46 & 29.37\\

    Full-grid + Visibility-E
    & & $\checkmark$ & $\checkmark$
    & 54.15 & 78.73&88.76 &2.17 & 16.32&30.82 & 33.79&26.93 &29.54 \\

    \midrule
    \textbf{CascadeJEPA}
    & $\checkmark$ & $\checkmark$ & $\checkmark$  
    & \textbf{56.31} & \textbf{79.87} & \textbf{89.51}
    & \textbf{4.26} & \textbf{25.17} & \textbf{39.42}
    & \textbf{45.67} & \textbf{32.39} & \textbf{33.59}  \\

    \bottomrule
\end{tabular}
\vspace{-1.0em}
\end{table*}

\begin{table*}[t]
\centering
\caption{Ablation studies on cross-attention placement and first-stage hyperparameters. Lin.: ImageNet-1K linear probing with 1\% labels. }
\vspace{-1em}
\label{tab:abla_hyper}
\footnotesize
\setlength{\tabcolsep}{4.5pt}
\renewcommand{\arraystretch}{1.12}
\begin{subtable}[t]{0.24\textwidth}
    \centering
    \caption{Masked target blocks}
    \label{tab:abla_patch_num}
    \begin{tabular}{cccc}
        \toprule
        Blocks & Lin. & $\delta_1$  & 5px \\
        \midrule
        1 & 2.59 & 34.06 & 1.41 \\
        2 & 41.91 & \textbf{56.31} & \textbf{4.26} \\
        3 & \textbf{48.46} & 55.01 & 3.78 \\
        \bottomrule
    \end{tabular}
\end{subtable}
\hfill
\begin{subtable}[t]{0.24\textwidth}
    \centering
    \caption{Target-block scale}
    \label{tab:abla_mask_ratio}
    \begin{tabular}{cccc}
        \toprule
        Block scale & Lin. & $\delta_1$  & 5px \\
        \midrule
        0.15--0.20 & 31.03 & 53.96 & 3.97 \\
        0.20--0.25 & 41.91 & \textbf{56.31} & \textbf{4.26} \\
        0.25--0.30 & \textbf{47.54} & 54.28 & 3.61 \\
        \bottomrule
    \end{tabular}
\end{subtable}
\hfill
\begin{subtable}[t]{0.23\textwidth}
    \centering
    \caption{Predictor depth}
    \label{tab:abla_depth}
    \begin{tabular}{cccc}
        \toprule
        Depth & Lin. & $\delta_1$  & 5px\\
        \midrule
        1 & 38.43 & 55.39 & 3.97 \\
        2 & 41.91 & \textbf{56.31} & \textbf{4.26} \\
        3 & 42.05 & 55.33 & 4.05 \\
        4 & \textbf{43.46} & 54.77 & 3.89 \\
        \bottomrule
    \end{tabular}
\end{subtable}
\hfill
\begin{subtable}[t]{0.23\textwidth}
    \centering
    \caption{Cross-attention placement}
    \label{tab:abla_cross_attn}
    \begin{tabular}{cccc}
        \toprule
        Layers & Lin. & $\delta_1$  & 5px \\
        \midrule
        1 & 41.91 & \textbf{56.31} & \textbf{4.26} \\
        2 & \textbf{44.49} & 55.54 & 4.01 \\
        4 & 39.06 & 55.58 & 3.80 \\
        6 & 38.78 & 53.87 & 2.35 \\
        \bottomrule
    \end{tabular}
\end{subtable}
\vspace{-1.0em}
\end{table*}


\textbf{Depth Estimation.}
To further evaluate the representation of 3D spatial structure, we perform monocular depth estimation on NAVI~\cite{jampani2023navi} using the same frozen-backbone probing protocol~\cite{el2024probing}. We report depth accuracy at three thresholds. Tab.~\ref{tab:normal_depth} shows CascadeJEPA substantially improves upon I-JEPA across all thresholds, with gains of 12.81, 10.03, and 6.29 percentage points, respectively. Notably, CascadeJEPA achieves the highest $\delta_1$ accuracy among all compared ImageNet-1K-pretrained methods, including invariance-driven baselines. These results indicate that the benefits of spatially coupled cascaded prediction extend beyond local geometric details to depth-related 3D structure. 

\textbf{ImageNet Classification.}
 We perform end-to-end supervised fine-tuning on ImageNet-1K~\cite{russakovsky2015imagenet} using a unified MAE-based MMPreTrain~\cite{2023mmpretrain}  implementation with identical hyperparameters. As shown in Tab.~\ref{tab:semantic_comparison}, CascadeJEPA achieves 83.50\% Top-1 accuracy with fewer pretraining epochs, outperforming I-JEPA and achieving comparable performance to MAE. These results demonstrate that CascadeJEPA enhances geometric representations while maintaining strong semantic recognition.


\textbf{Qualitative Evaluation.}
\label{sec:qualitative}
As illustrated in Fig.~\ref{fig:vis_task}, we qualitatively compare the geometric representations learned by I-JEPA and CascadeJEPA. CascadeJEPA produces more spatially coherent depth predictions with clearer object boundaries, more consistent surface normal estimates that capture fine-grained geometric structures, and dense features exhibiting more distinct spatial organization.  



\subsection{Ablation Studies}
\label{sec:ablation}

\textbf{Component-wise Ablation.}
Tab.~\ref{tab:component_ablation} evaluates three design choices:
cascaded prediction with cross-attention (Cascade),
full-grid context encoding (Full-grid), and visibility-enhanced
masking (Visibility-E). All ablation variants are pretrained for 300 epochs. The Full-grid + Visibility-E retains
two prediction stages but removes cross-attention.

\textit{1) Depth estimation benefits from richer visible context.}
Visibility-E improves depth accuracy from
51.10\% to 55.12\% over the cascade baseline, while removing it from CascadeJEPA
causes a 3.57 drop. This suggests that richer visible
context provides more comprehensive spatial cues for
recovering single-image 3D structure.

\textit{2) Matching benefits from both full-grid spatial context and
cross-attention.}
For semantic matching, introducing cross-attention alone
improves $d=0$ from 25.65\% to 29.67\%.
This suggests that cross-crop interactions facilitate
semantic representation learning, aligning with recent
findings~\cite{sunagad2026two}.
Visibility-E further improves semantic
matching, as more comprehensive spatial cues
help establish semantically consistent correspondences
across views.

For geometric matching, both Full-grid and Visibility-E
improve correspondence accuracy when incorporated into
the cascade baseline. Notably, integrating all three
components further increases Recall@5px to 4.26\%, substantially outperforming the variant without cross-attention (2.17\%).
These results suggest that cross-attention can more effectively facilitate fine-grained geometric correspondence when supported by sufficiently rich spatial context.

We hypothesize that differently scaled crops of the same
image provide complementary patch-level features
at varying spatial granularities. Full-grid encoding
preserves spatial layout information, while Visibility-E
enriches contextual details. Building on this, cross-attention facilitates
multi-granularity patch-level interactions across stages. Such multi-granularity feature interaction is helpful for 3D geometric representation learning.

\begin{figure*}[t]  
  \centering
    \includegraphics[width=1.0\linewidth]{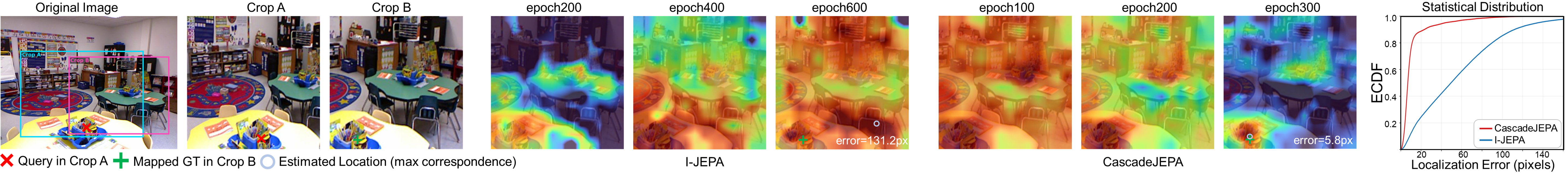} 
    \vspace{-2.0em}
    \caption{Cross-crop spatial correspondence across pretraining epochs on NYUv2, with response maps and localization error distributions.}  
    \label{fig:vis_attn}    
    \vspace{-1.0em}
\end{figure*}

\textbf{Hyperparameter Analysis.}
In CascadeJEPA, the second stage follows the original I-JEPA configuration, with a predictor depth of 6, a target-block scale of 0.15--0.20, and 4 masked target blocks. Tab.~\ref{tab:abla_hyper} presents ablation studies on cross-attention placement and first-stage hyperparameters.

\textit{1) Masked Target Blocks.}
Reducing masked target blocks from 4 to 2 improves fine-grained geometric representations, highlighting the importance of sufficient contextual visibility. However, using only one block severely degrades performance, potentially undermining effective latent representation learning during pretraining.

\textit{2) Target-Block Scale.}
Increasing the target-block scale from 0.15–0.20 to 0.20–0.25 improves geometric performance, suggesting a balance between contextual visibility and prediction difficulty. However, further increasing the scale to 0.25–0.30 leads to degraded geometric representations, potentially due to reduced contextual visibility.


\textit{3) Predictor Depth.}
With richer visible spatial context, a shallow two-layer predictor yields the strongest geometric accuracy, suggesting that sufficient contextual information reduces the need for deeper latent prediction, which may favor semantic abstraction over fine-grained spatial details.

\textit{4) Cross-attention Placement.}
Inserting cross-attention before  the first predictor layer outperforms later placements, suggesting that early spatial coupling allows subsequent layers to better integrate first-stage contextual information.

\subsection{Discussion}
\label{sec:discussion}


\textbf{Understanding CascadeJEPA's Geometric Improvements.}
To investigate why CascadeJEPA improves geometric representations, we analyze the evolution of cross-crop feature correspondences during pretraining using two overlapping crops. As shown in Fig.~\ref{fig:vis_attn}, I-JEPA exhibits increasingly diffuse correspondence responses as pretraining progresses, whereas CascadeJEPA develops more spatially concentrated responses around the ground-truth location. At epoch 300, CascadeJEPA achieves a localization error of only 5.8 pixels, compared with 131.2 pixels for I-JEPA. Furthermore, the empirical cumulative distribution functions (ECDFs) over the NYUv2 test set show consistently lower localization errors for CascadeJEPA. We hypothesize that spatially coupled cascaded prediction promotes richer patch-level interactions, allowing CascadeJEPA to better capture inter-patch spatial relationships than I-JEPA. These interactions may encourage spatially discriminative representations that preserve fine-grained geometric distinctions, consistent with the improved cross-crop correspondences.




\begin{table}[t]
\centering
\caption{Feature forecasting evaluation with DINO-Foresight~\cite{karypidis2026dino}.}
\vspace{-1.0em}
\label{tab:temporal_prediction}
\footnotesize
\setlength{\tabcolsep}{6pt}
\renewcommand{\arraystretch}{1.12}
\begin{tabular}{lcccc}
    \toprule
    Method & Copy Last & Prediction & Oracle & Pred. Rate \\
    \midrule
    I-JEPA~\cite{assran2023ijepa}
    &  32.46  & 37.87 & 45.61 & 0.411 \\
    CascadeJEPA
    & \textbf{33.55} & \textbf{40.41}
    & \textbf{47.99} & \textbf{0.475} \\
    \bottomrule
\vspace{-2.5em}
\end{tabular}
\end{table}
\begin{table}[t]
\centering
\caption{World model planning evaluation with DINO-WM~\cite{zhou2024dino}.}
\vspace{-1.0em}
\label{tab:ijepa_cascadejepa}
\footnotesize
\setlength{\tabcolsep}{8.5pt}
\renewcommand{\arraystretch}{1.12}
\begin{tabular}{lcccc}
    \toprule
    Method & PointMaze & Wall & PushT & Average \\
    \midrule
    I-JEPA~\cite{assran2023ijepa}
    & 0.773 & 0.287 & 0.833 & 0.631 \\
    CascadeJEPA
    & \textbf{0.960} & \textbf{0.420}
    & \textbf{0.840} & \textbf{0.740} \\
    \bottomrule
\end{tabular}
\vspace{-1.0em}
\end{table}

\textbf{Future Feature Forecasting.}
To investigate whether enhanced geometric representations benefit future visual understanding, we adopt DINO-Foresight~\cite{karypidis2026dino}, replacing its DINOv2 encoder with I-JEPA and CascadeJEPA while keeping  other designs unchanged. Following its evaluation protocol, we assess future semantic segmentation using mIoU. As shown in Tab.~\ref{tab:temporal_prediction}, CascadeJEPA improves forecasting mIoU from 37.87\% to 40.41\%, while increasing the normalized prediction rate from 0.411 to 0.475. These improvements are consistent with the enhanced geometric representations observed in our evaluations, suggesting their potential relevance to future feature forecasting.

\begin{figure}[t]  
  \centering
    \includegraphics[width=1.0\linewidth]{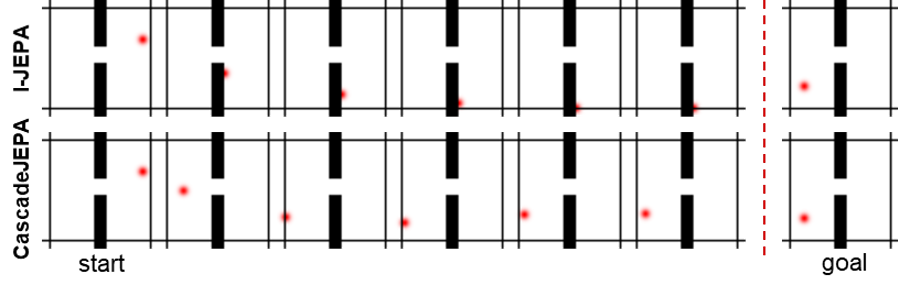} 
    \vspace{-2.0em}
    \caption{Plans generated by I-JEPA and CascadeJEPA on Wall.}  
    \label{fig:vis_wm}   
     \vspace{-1.0em}
\end{figure}

\textbf{World Model Planning.}  We further investigate whether enhanced geometric representations benefit action planning using DINO-WM~\cite{zhou2024dino}. We replace its original DINOv2 encoder with I-JEPA and CascadeJEPA while keeping the architecture unchanged. As shown in Tab.~\ref{tab:ijepa_cascadejepa}, CascadeJEPA improves the average planning success rate from 63.1\% to 74.0\%, with particularly notable gains on PointMaze and Wall. Furthermore, Fig.~\ref{fig:vis_wm} visualizes the generated plans on Wall, where CascadeJEPA produces trajectories that better navigate around obstacles toward the goal, whereas I-JEPA exhibits less effective spatial planning. These results provide evidence that geometrically informative representations can benefit downstream   action-conditioned planning.

\textbf{Limitations.}
CascadeJEPA incurs additional pretraining overhead due to its two-stage prediction framework. Moreover, as with the original I-JEPA, its performance may be sensitive to hyperparameter configurations.


\section{Conclusions}
\label{sec:conclusion}

Our findings reveal that improved semantic learning in I-JEPA does not necessarily translate into better geometric representations, suggesting potential limitations of conventional single-stage latent prediction in capturing fine-grained geometric relationships. Building on this insight, CascadeJEPA demonstrates that spatially coupled cascaded prediction can substantially enhance 3D geometric representations with minimal architectural modifications to I-JEPA, while maintaining strong semantic performance. Evaluations further show that these improvements extend beyond geometric perception to future feature forecasting and world model planning. These results highlight the potential of geometrically informative representations for developing more effective predictive world models.


\section*{LLM Usage Statement}
ChatGPT was employed as an auxiliary tool for implementing and debugging experimental scripts, as well as refining the manuscript's language and presentation. The research concepts, methodological contributions, experimental interpretations, and conclusions are entirely the authors' own. All AI-assisted materials were independently examined and validated by the authors, who assume full responsibility for the final manuscript.

{
    \small
    \bibliographystyle{ieeenat_fullname}
    \bibliography{main}

@misc{lecun2022path,
 author={Yann LeCun},
 title={A Path Towards Autonomous Machine Intelligence},
 year={2022},
 howpublished={OpenReview},
 url={https://openreview.net/forum?id=BZ5a1r-kVsf}
}

@inproceedings{assran2023ijepa,
 author={Mahmoud Assran and Quentin Duval and Ishan Misra and Piotr Bojanowski and Pascal Vincent and Michael Rabbat and Yann LeCun and Nicolas Ballas},
 title={Self-Supervised Learning from Images with a Joint-Embedding Predictive Architecture},
 booktitle={Proceedings of the IEEE/CVF Conference on Computer Vision and Pattern Recognition},
 pages={15619--15629}, year={2023},
 url={https://openaccess.thecvf.com/content/CVPR2023/html/Assran_Self-Supervised_Learning_From_Images_With_a_Joint-Embedding_Predictive_Architecture_CVPR_2023_paper.html}
}

@inproceedings{zhou2026understanding,
  title={Understanding Geometric Representations in Self-Supervised Vision Transformers via Subspace Intervention},
  author={Weichen Zhou and Yawen Zou and Chunzhi Gu and Ran Dong and Haoran Xie and Chao Zhang},
  booktitle={Proceedings of the European Conference on Computer Vision},
  year={2026},
  url={https://github.com/Zhou-Weichen/Geosubprobe}
}

@inproceedings{dosovitskiy2020image,
 author={Alexey Dosovitskiy and Lucas Beyer and Alexander Kolesnikov and Dirk Weissenborn and Xiaohua Zhai and Thomas Unterthiner and Mostafa Dehghani and Matthias Minderer and Georg Heigold and Sylvain Gelly and Jakob Uszkoreit and Neil Houlsby},
 title={An Image is Worth 16x16 Words: Transformers for Image Recognition at Scale},
 booktitle={International Conference on Learning Representations}, year={2021},
 url={https://arxiv.org/abs/2010.11929}
}

@inproceedings{touvron2021training,
  title={Training Data-Efficient Image Transformers \& Distillation through Attention},
  author={Hugo Touvron and Matthieu Cord and Matthijs Douze and Francisco Massa and Alexandre Sablayrolles and Herv{\'e} J{\'e}gou},
  booktitle={Proceedings of the 38th International Conference on Machine Learning},
  series={Proceedings of Machine Learning Research},
  volume={139},
  pages={10347--10357},
  year={2021},
  publisher={PMLR},
  url={https://proceedings.mlr.press/v139/touvron21a.html}
}

@inproceedings{touvron2022deit,
  title={{DeiT III}: Revenge of the {ViT}},
  author={Hugo Touvron and Matthieu Cord and Herv{\'e} J{\'e}gou},
  booktitle={Proceedings of the European Conference on Computer Vision},
  pages={516--533},
  year={2022},
  url={https://arxiv.org/abs/2204.07118}
}

@inproceedings{chen2021empirical,
 author={Xinlei Chen and Saining Xie and Kaiming He},
 title={An Empirical Study of Training Self-Supervised Vision Transformers},
 booktitle={Proceedings of the IEEE/CVF International Conference on Computer Vision},
 pages={9640--9649}, year={2021},
 url={https://openaccess.thecvf.com/content/ICCV2021/html/Chen_An_Empirical_Study_of_Training_Self-Supervised_Vision_Transformers_ICCV_2021_paper.html}
}

@inproceedings{he2022masked,
 author={Kaiming He and Xinlei Chen and Saining Xie and Yanghao Li and Piotr Doll{\'a}r and Ross Girshick},
 title={Masked Autoencoders Are Scalable Vision Learners},
 booktitle={Proceedings of the IEEE/CVF Conference on Computer Vision and Pattern Recognition},
 pages={16000--16009}, year={2022},
 url={https://openaccess.thecvf.com/content/CVPR2022/html/He_Masked_Autoencoders_Are_Scalable_Vision_Learners_CVPR_2022_paper.html}
}

@inproceedings{caron2021emerging,
 author={Mathilde Caron and Hugo Touvron and Ishan Misra and Herv{\'e} J{\'e}gou and Julien Mairal and Piotr Bojanowski and Armand Joulin},
 title={Emerging Properties in Self-Supervised Vision Transformers},
 booktitle={Proceedings of the IEEE/CVF International Conference on Computer Vision},
 pages={9650--9660}, year={2021},
 url={https://openaccess.thecvf.com/content/ICCV2021/html/Caron_Emerging_Properties_in_Self-Supervised_Vision_Transformers_ICCV_2021_paper.html}
}

@inproceedings{zhou2022ibot,
 author={Jinghao Zhou and Chen Wei and Huiyu Wang and Wei Shen and Cihang Xie and Alan Yuille and Tao Kong},
 title={{iBOT}: Image {BERT} Pre-Training with Online Tokenizer},
 booktitle={International Conference on Learning Representations}, year={2022},
 url={https://github.com/bytedance/ibot}
}

@inproceedings{he2026dseqjepa,
  title={{DSeq-JEPA}: Discriminative Sequential Joint-Embedding Predictive Architecture},
  author={Xiangteng He and Shunsuke Sakai and Shivam Chandhok and Sara Beery and Kun Yuan and Nicolas Padoy and Tatsuhito Hasegawa and Leonid Sigal},
  booktitle={Proceedings of the European Conference on Computer Vision},
  year={2026},
  url={https://dseqjepa-project.com/}
}

@article{xu2025nepa,
 author={Sihan Xu and Ziqiao Ma and Wenhao Chai and Xuweiyi Chen and Weiyang Jin and Joyce Chai and Saining Xie and Stella X. Yu},
 title={Next-Embedding Prediction Makes Strong Vision Learners},
 journal={arXiv preprint arXiv:2512.16922}, year={2025},
 url={https://arxiv.org/abs/2512.16922}
}

@article{balestriero2025lejepa,
 author={Randall Balestriero and Yann LeCun},
 title={{LeJEPA}: Provable and Scalable Self-Supervised Learning without the Heuristics},
 journal={arXiv preprint arXiv:2511.08544}, year={2025},
 url={https://arxiv.org/abs/2511.08544}
}

@article{bardes2024vjepa,
 author={Adrien Bardes and Quentin Garrido and Jean Ponce and Xinlei Chen and Michael Rabbat and Yann LeCun and Mahmoud Assran and Nicolas Ballas},
 title={Revisiting Feature Prediction for Learning Visual Representations from Video},
 journal={Transactions on Machine Learning Research}, year={2024},
 url={https://arxiv.org/abs/2404.08471}
}

@article{assran2025vjepa2,
 author={Mido Assran and Adrien Bardes and David Fan and Quentin Garrido and Russell Howes and Mojtaba Komeili and Matthew Muckley and Ammar Rizvi and Claire Roberts and Koustuv Sinha and Artem Zholus and Sergio Arnaud and Abha Gejji and Ada Martin and Francois Robert Hogan and Daniel Dugas and Piotr Bojanowski and Vasil Khalidov and Patrick Labatut and Francisco Massa and Marc Szafraniec and Kapil Krishnakumar and Yong Li and Xiaodong Ma and Sarath Chandar and Franziska Meier and Yann LeCun and Michael Rabbat and Nicolas Ballas},
 title={{V-JEPA 2}: Self-Supervised Video Models Enable Understanding, Prediction and Planning},
 journal={arXiv preprint arXiv:2506.09985}, year={2025},
 url={https://arxiv.org/abs/2506.09985}
}

@article{murlabadia2026vjepa21,
 author={Lorenzo Mur-Labadia and Matthew Muckley and Amir Bar and Mido Assran and Koustuv Sinha and Mike Rabbat and Yann LeCun and Nicolas Ballas and Adrien Bardes},
 title={{V-JEPA 2.1}: Unlocking Dense Features in Video Self-Supervised Learning},
 journal={arXiv preprint arXiv:2603.14482}, year={2026},
 url={https://arxiv.org/abs/2603.14482}
}

@article{kuhn2026levjepa,
 author={Lukas Kuhn and Lucas Maes and Giuseppe Serra and Quentin Le Lidec and Yann LeCun and Randall Balestriero and Florian Buettner},
 title={{LeVJEPA}: Efficient \& Scalable Video Pretraining without the Heuristics},
 journal={arXiv preprint arXiv:2608.27395}, year={2026},
 url={https://arxiv.org/abs/2608.27395}
}

@inproceedings{jampani2023navi,
  title={{NAVI}: Category-Agnostic Image Collections with High-Quality {3D} Shape and Pose Annotations},
  author={Varun Jampani and Kevis-Kokitsi Maninis and Andreas Engelhardt and Arjun Karpur and Karen Truong and Kyle Sargent and Stefan Popov and Andr{\'e} Araujo and Ricardo Martin-Brualla and Kaushal Patel and Daniel Vlasic and Vittorio Ferrari and Ameesh Makadia and Ce Liu and Yuanzhen Li and Howard Zhou},
  booktitle={Advances in Neural Information Processing Systems},
  volume={36},
  pages={76061--76084},
  year={2023},
  doi={10.52202/075280-3323},
  url={https://proceedings.neurips.cc/paper_files/paper/2023/hash/efc90033e6e1b05485312dd09fe302b8-Abstract-Datasets_and_Benchmarks.html}
}

@inproceedings{silberman2012indoor,
  title={Indoor Segmentation and Support Inference from {RGBD} Images},
  author={Nathan Silberman and Derek Hoiem and Pushmeet Kohli and Rob Fergus},
  booktitle={Proceedings of the European Conference on Computer Vision},
  pages={746--760},
  year={2012},
  url={https://cs.nyu.edu/~fergus/datasets/nyu_depth_v2.html}
}

@inproceedings{dai2017scannet,
 author={Angela Dai and Angel X. Chang and Manolis Savva and Maciej Halber and Thomas Funkhouser and Matthias Nie{\ss}ner},
 title={{ScanNet}: Richly-Annotated {3D} Reconstructions of Indoor Scenes},
 booktitle={Proceedings of the IEEE Conference on Computer Vision and Pattern Recognition},
 pages={2432--2443}, year={2017},
 url={https://openaccess.thecvf.com/content_cvpr_2017/html/Dai_ScanNet_Richly-Annotated_3D_CVPR_2017_paper.html}
}

@article{min2019spair,
 author={Juhong Min and Jongmin Lee and Jean Ponce and Minsu Cho},
 title={{SPair-71k}: A Large-Scale Benchmark for Semantic Correspondence},
 journal={arXiv preprint arXiv:1908.10543}, year={2019},
 url={https://cvlab.postech.ac.kr/research/SPair-71k/}
}

@article{russakovsky2015imagenet,
  title={{ImageNet} Large Scale Visual Recognition Challenge},
  author={Olga Russakovsky and Jia Deng and Hao Su and Jonathan Krause and Sanjeev Satheesh and Sean Ma and Zhiheng Huang and Andrej Karpathy and Aditya Khosla and Michael Bernstein and Alexander C. Berg and Li Fei-Fei},
  journal={International Journal of Computer Vision},
  volume={115},
  number={3},
  pages={211--252},
  year={2015},
  doi={10.1007/s11263-015-0816-y},
  url={https://doi.org/10.1007/s11263-015-0816-y}
}

@article{goodale1992separate,
  title={Separate Visual Pathways for Perception and Action},
  author={Goodale, Melvyn A. and Milner, A. David},
  journal={Trends in Neurosciences},
  volume={15},
  number={1},
  pages={20--25},
  year={1992},
  publisher={Elsevier}
}

@inproceedings{chen2025probing,
  title={Probing the Mid-level Vision Capabilities of Self-Supervised Learning},
  author={Chen, Xuweiyi and Marks, Markus and Cheng, Zezhou},
  booktitle={Proceedings of the IEEE/CVF Conference on Computer Vision and Pattern Recognition},
  pages={30095--30105},
  year={2025}
}

@inproceedings{el2024probing,
  title={Probing the {3D} Awareness of Visual Foundation Models},
  author={El Banani, Mohamed and Raj, Amit and Maninis, Kevis-Kokitsi and Kar, Abhishek and Li, Yuanzhen and Rubinstein, Michael and Sun, Deqing and Guibas, Leonidas and Johnson, Justin and Jampani, Varun},
  booktitle={Proceedings of the IEEE/CVF Conference on Computer Vision and Pattern Recognition},
  pages={21795--21806},
  year={2024},
  url={https://openaccess.thecvf.com/content/CVPR2024/html/Banani_Probing_the_3D_Awareness_of_Visual_Foundation_Models_CVPR_2024_paper.html}
}

@inproceedings{karypidis2026dino,
  title={{DINO-Foresight}: Looking into the Future with {DINO}},
  author={Karypidis, Efstathios and Kakogeorgiou, Ioannis and Gidaris, Spyridon and Komodakis, Nikos},
  booktitle={Advances in Neural Information Processing Systems},
  volume={38},
  pages={163779--163811},
  year={2025},
  doi={10.52202/085713-5466}
}

@inproceedings{zhou2024dino,
  title={{DINO-WM}: World Models on Pre-trained Visual Features Enable Zero-shot Planning},
  author={Zhou, Gaoyue and Pan, Hengkai and LeCun, Yann and Pinto, Lerrel},
  booktitle={Proceedings of the 42nd International Conference on Machine Learning},
  series={Proceedings of Machine Learning Research},
  volume={267},
  pages={79115--79135},
  year={2025},
  publisher={PMLR},
  url={https://proceedings.mlr.press/v267/zhou25t.html}
}

@misc{2023mmpretrain,
  title={{OpenMMLab}'s Pre-training Toolbox and Benchmark},
  author={{MMPreTrain Contributors}},
  howpublished={\url{https://github.com/open-mmlab/mmpretrain}},
  year={2023}
}

@article{sunagad2026two,
  title={Two Global Crops Suffice: Locating Semantic Emergence in DINO-Style Self-Supervised Learning},
  author={Sunagad, Basavaraj and Jesslen, Artur and Kortylewski, Adam},
  journal={arXiv preprint arXiv:2609.28187},
  year={2026}
}

@article{connor2017integration,
  title={Integration of Objects and Space in Perception and Memory},
  author={Connor, Charles E. and Knierim, James J.},
  journal={Nature Neuroscience},
  volume={20},
  number={11},
  pages={1493--1503},
  year={2017},
  doi={10.1038/nn.4657},
  publisher={Nature Publishing Group}
}
}


\end{document}